\pdfoutput=1

\documentclass[11pt]{article}

\usepackage{tikz}
\usepackage{booktabs}
\usetikzlibrary{shapes,arrows,positioning,fit,backgrounds}
\usepackage{xcolor}
\usepackage{amsmath}
\usepackage{enumitem}
\usepackage[preprint]{acl}

\usepackage{times}
\usepackage{latexsym}

\usepackage[T1]{fontenc}

\usepackage[utf8]{inputenc}

\usepackage{microtype}

\usepackage{inconsolata}

\usepackage{graphicx}

\title{Mind the Gap: Assessing Wiktionary’s Crowd-Sourced Linguistic Knowledge on Morphological Gaps in Two Related Languages}

\author{Jonathan Sakunkoo \\
  Stanford University OHS \\
  \texttt{jonkoo@ohs.stanford.edu} \\\And
  Annabella Sakunkoo \\
  Stanford University OHS \\
\texttt{apianist@ohs.stanford.edu} \\}

\begin{document}
\maketitle
\begin{abstract}
 Morphological defectivity is an intriguing and understudied phenomenon in linguistics. Addressing defectivity, where expected inflectional forms are absent, is essential for improving the accuracy of NLP tools in morphologically rich languages. However, traditional linguistic resources often lack coverage of morphological gaps as such knowledge requires significant human expertise and effort to document and verify. For scarce linguistic phenomena in under-explored languages, Wikipedia and Wiktionary often serve as among the few accessible resources. Despite their extensive reach, their reliability has been a subject of controversy. This study customizes a novel neural morphological analyzer to annotate Latin and Italian corpora. Using the massive annotated data, crowd-sourced lists of defective verbs compiled from Wiktionary are validated computationally. Our results indicate that while Wiktionary provides a highly reliable account of  Italian morphological gaps, 7\% of Latin lemmata listed as defective show strong corpus evidence of being non-defective. This discrepancy highlights  potential limitations of crowd-sourced wikis as definitive sources of linguistic knowledge, particularly for less-studied phenomena and languages, despite their value as resources for rare linguistic features. By providing scalable tools and methods for quality assurance of crowd-sourced data, this work advances computational morphology and expands linguistic knowledge of defectivity in non-English, morphologically rich languages. 

\end{abstract}

\section{Introduction}


\begin{quote}
    The past tense of "forgo" is forwent.
So, you would say: "I forwent this position."
It's a bit formal or uncommon in modern usage, but grammatically correct.
\end{quote}
Above is a response from GPT-4o when asked what the past tense for “forgo” is. Similarly, Llama 3.2 confidently replies that \begin{quote}The past tense of "forgo" is "forwent".
\end{quote} 
Yet, most English speakers would find \emph{forwent} ineffable \cite{Gorman2023} and unacceptable \cite{EmbickMarantz2008}. Most English speakers are actually unable to find the right, natural form for the past tense of \emph{forgo} \cite{Gorman2019}. Similarly, \emph{beware} functions exclusively as a positive imperative (e.g. beware the bear!), and \emph{BEGO} can only appear as the imperative \emph{begone!}
Words such as these are instances of defective verbs or morphological gaps in which expected forms are missing—a problematic intrusion of morphological idiosyncrasy \cite{BaermanCorbett2010}. In other words, a lexeme is defective if at least one of its possible inflectional variants is ineffable \cite{Gorman2023} or exhibits relative non-use \cite{Sims2006minding}.

In Latin, \emph{ai\=o} `to speak' lacks the first- and second-person plural present forms. Another defective verb is \emph{inquam} `to say', also restricted to an incomplete subset of forms, such as the third person singular in the present and perfect indicative (e.g. \emph{inquit}) \cite{Oniga2014}. 

While inflectional gaps are not a recent discovery, they "remain poorly understood" \cite{BaermanCorbett2010}. Since NLP systems often assume regular paradigms, recognizing and accounting for defectivity would improve the accuracy so as to not use or suggest forms that do not exist, especially for less-studied and morphologically rich languages where inflectional gaps are more common. \citet{gorman2024acquiring} applied UDTube to discriminate defective from non-defective words in Russian and Greek. While curated lists of defective verbs exist for languages such as Russian and Greek, trusted and verified resources remain scarce for many others, including Latin and Italian. 
For scarce linguistic phenomena in less-studied languages, Wikipedia and Wiktionary often serve as two of the few widely accessible and frequently utilized resources, consistently ranked among the most popular websites globally, attracting over 4.5 billion unique monthly visitors. With its extensive reach and usage, crowd-sourced content is a potentially valuable but underexplored resource; recognizing this, projects like UniMorph \cite{kirov2018unimorph} have made  progress in extracting morphological data from Wiktionary. However, despite its many virtues, Wiktionary (along with the rest of the Wikimedia family)'s user-contributed nature has sparked controversy on  trustworthiness, and the reliability and completeness of Wiktionary still present challenges. 

In this study, we conduct computational analyses of inflectional gaps by customizing UDTube \citep{yakubov2024how}\footnote{\url{https://github.com/CUNY-CL/udtube}}, a scalable state-of-the-art neural morphological analyzer trained with Universal Dependencies (a collection of corpora of morphologically annotated text in different languages), to incorporate mBERT \cite{devlin-etal-2019-bert} as an encoder. We apply this enhanced model to annotate large corpora of text in Latin (640MB, 390 million words) and Italian (8.3GB, 5 billion words). The resulting massive annotated data are then used to validate lists of defective verbs scraped and compiled from Wiktionary's Latin and Italian pages to verify which verbs are confirmed computationally to be defective or non-defective. 

We model defectivity after how children might learn what the gaps or defective forms are--in other words, learn what is missing. \citet{brown1970derivational} showed that parents typically provide explicit feedback on the truth value of a child's articulation but rarely correct grammatical errors, such as inflection, thus implying that children do not acquire morphology through explicit negative evidence. Similarly, \citet{baronian2005north} reinforced the idea that morphological gaps are not taught directly. While the exact process by which children acquire defectivity remains unclear, many scholars in linguistics and language learning agree that gaps are primarily learned through \textbf{indirect} (or implicit) \textbf{negative evidence} (INE) (\citet{orgun1999mparse}; \citet{johansson1999learning}; \citet{Sims2006minding}). 
By comparing Wiktionary's defectivity lists against this INE-based computational approach, we evaluate whether crowd-sourced linguistic knowledge aligns with human language learning patterns.

Our findings indicate that nearly 80\% of inflectional gaps in Italian and 70\% in Latin listed in Wiktionary strongly align with our computational INE results while 4\% of Italian and 7\% of Latin lemmata labeled as defective in Wiktionary show a high tendency to actually be non-defective, thus suggesting a degree of reliability in Wiktionary's linguistic data, despite coming from unreferenced, user-generated sources. The study also identifies multiple inaccuracies, particularly in Latin, and highlights the need for more rigorous expert verification in crowd-sourced linguistic resources.

This study examines the potential and limitations of crowd-sourced content as a supplementary linguistic resource, based on a model of how children learn defectivity through indirect negative evidence. By using a novel, scalable approach for computationally analyzing morphological gaps, it advances the intersection of computational methods and linguistics as it contributes to quality assurance of crowdsourced content and addresses gaps in linguistic knowledge. 

\begin{figure*}[!t]
\centering
\begin{tikzpicture}[node distance=2.2cm, auto, scale=0.75, transform shape] 
    \tikzstyle{block1} = [rectangle, draw, fill=blue!20, 
                         text width=3.2cm, text centered, 
                         rounded corners, minimum height=1.1cm] 
    \tikzstyle{block2} = [rectangle, draw, fill=cyan!20, 
                         text width=3.2cm, text centered, 
                         rounded corners, minimum height=1.1cm] 
    \tikzstyle{block3} = [rectangle, draw, fill=orange!20, 
                         text width=3.2cm, text centered, 
                         rounded corners, minimum height=1.1cm] 
    \tikzstyle{line} = [draw, -latex', thick]
    
    \node [block1] (ud) {Universal Dependencies\\(Annotated Corpus)};
    \node [block1, right=of ud] (udtube) {UDTube\\(with mBERT)};
    \node [block1, right=of udtube] (analyzer) {Trained\\Morphological Analyzer};
    
    \node [block2, below=1.6cm of ud] (cc) {Common Crawl\\(Raw Text Corpus)};
    \node [block2, right=of cc] (tokenize) {Tokenize with\\UDPipe};
    \node [block2, right=of tokenize] (tagged) {Morphologically\\Tagged Corpus};
    
    \node [block3, below=1.6cm of cc] (wiki) {Wiktionary\\(Defective Forms)};
    \node [block3, below=1.6cm of tagged] (valid) {Attestation\\Analysis};
    
    \path [line] (ud) -- (udtube);
    \path [line] (udtube) -- (analyzer);
    \path [line] (cc) -- (tokenize);
    \path [line] (tokenize) -- (analyzer);
    \path [line] (analyzer) -- (tagged);
    \path [line] (wiki) -- (valid);
    \path [line] (tagged) -- (valid);
    
    \node [above=0.3cm of ud] {\textbf{Training Phase}};
    \node [above=0.3cm of cc] {\textbf{Analysis Phase}};
    \node [above=0.3cm of wiki] {\textbf{Validation Phase}};
\end{tikzpicture}
\caption{Workflow for computational validation of morphological gaps, using UDTube}
\label{fig:workflow}
\end{figure*}
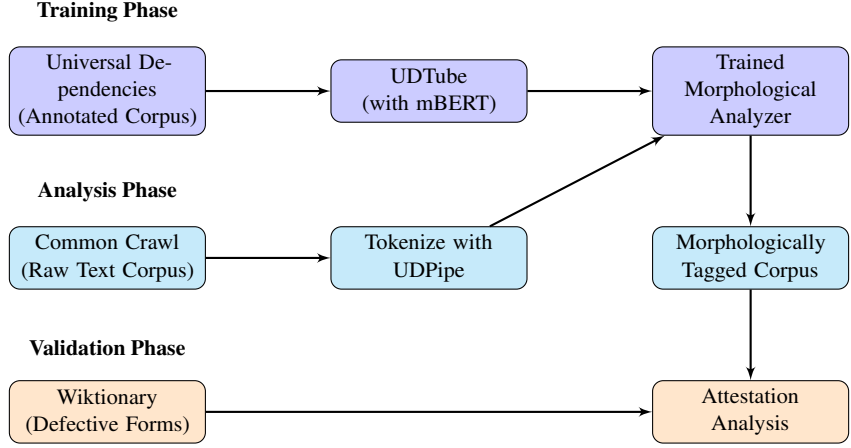

\section{Data}

We employ the following data sources in the computational validation of morphological gaps.


\textbf{Universal Dependencies (UD)}  \cite{nivre-etal-2017-universal}:  We utilize two of the largest available Latin and Italian treebanks—UD Latin ITTB and UD Italian VIT—to train our morphological analyzer.


\textbf{Common Crawl (CC-100)} \cite{wenzek-etal-2020-ccnet}: From CC-100, we use an 8.3GB 
dataset containing 5B tokens of Italian text and a 640MB dataset with over 390M tokens of Latin text. 



\textbf{Wiktionary:} We scrape and compile lists of defective verbs and inflectional gaps from Latin and Italian pages of Wikitionary. This study focuses on Latin and Italian because of their reasonably large number of inflectional gaps and their representation in Wiktionary, which contains the most extensive lists of morphological gaps for these languages.



\section{Methodology}

As shown in Figure 1, this study uses a computational approach to validate inflectional gaps in Latin and Italian in three major steps:

\textbf{Training UDTube with UD:} As a neural morphological analyzer, UDTube's primary purpose is to decompose words morphologically and identify their morphological features. We trained UDTube using the mBERT encoder, a multilingual BERT model trained on 104 languages \cite{devlin-etal-2019-bert}, on the UD Italian and Latin treebanks. UDTube has been demonstrated to have superior performance in recent comparative studies \cite{yakubov2024how}, which show that it achieves high accuracy in morphological annotations, outperforming the popular UDPipe \cite{straka-etal-2016-udpipe} in multiple languages. Our tuned UDTube model has 98\% and 96\% accuracies in Features Morphological Annotations in Latin and Italian, respectively.

In hyperparameter tuning, optimal hyperparameters were determined using Weights and Biases, a tool for tracking and visualizing experiments. This step ensured that UDTube's configuration was fine-tuned for Latin and Italian datasets.

\textbf{Annotating Large-Scale Text:} The trained UDTube model is used to annotate text from the Common Crawl corpora. The process involved:

\begin{itemize}
    \item \textbf{Text Preprocessing:} The raw text was cleaned and tokenized using UDPipe \cite{straka-etal-2016-udpipe} into words.
    \item \textbf{Morphological Tagging:} Each token was analyzed and annotated with its lemma and morphological features, using the trained UDTube model. This produced a morphologically tagged corpus in CoNLL-U format.
    \item \textbf{Frequency Database:} From the tagged data, we generated a frequency database containing the occurrence counts for each morphological form of every lemma.
\end{itemize}

\textbf{Validating Defective Forms:} To verify the defective forms listed in Wiktionary, we applied the principle of \textbf{Indirect} (\textbf{Implicit}) \textbf{Negative Evidence} \citep{Gorman2019, BoydGoldberg2011}, a key mechanism in language acquisition by which learners infer defectivity: if a certain morphological form is defective, then it should not occur or occur extremely infrequently in usage. We employ two models to quantify the likelihood of non-defectivity. The first is \textbf{absolute frequency}. If a possible word has a high absolute frequency, it is unlikely to be defective. The second is \textbf{divergence from expected frequency}. If the frequency of a possible inflected word is significantly higher than expected, assuming all else is equal, it is unlikely to be defective.

For each attested inflected word $w$, there exist a corresponding lemma $l$ and a morphosyntactic feature bundle $f$. Let $p_w$, $p_l$, and $p_f$ denote the probability of a word, lemma, and feature bundle, respectively, calculated from maximum likelihood estimation using corpus frequencies. Assuming independence and all else equal, $p_w$ should be in proportion to $p_l \cdot p_f$. To measure \textbf{divergence from expected frequency}, how far a given inflected word has diverged from its expected probability, we use the \textbf{log-odds ratio} \cite{gorman2024acquiring}. 

\begin{center}
    Log-odds ratio: $L_w=\log\left({\dfrac{p_w}{p_l \cdot p_f}}\right)$
\end{center}


The log-odds ratio has been found to be the best surprisal or unexpectedness predictor for acceptability judgments \cite{lu-etal-2024-syntactic}. A log-odds ratio of 1.9 or higher is considered to indicate a large divergence \cite{Chen2010}.

The reliability of Wiktionary's crowd-sourced data was assessed by calculating the percentage of purported defective forms that aligned with our computational findings. The evaluation was grouped into true positives, which are cases where the Wiktionary-listed defective form was confirmed as absent or extremely rare in the corpus, and false positives, which are cases where a supposedly defective form was frequently attested in the corpus, indicating an error in Wiktionary. For discrepancies, we conducted manual reviews to determine whether they arose from corpus limitations, UDTube errors, or inaccuracies in Wiktionary.

\section{Results}

In evaluating defective lemmata listed in Wiktionary against corpus evidence, lemmata are classified into four groups:
      
      \textbf{Not Attested:} No inflected form of the lemma appears in the corpus, so we cannot confidently verify whether it is defective or not. These lemmata are excluded from our analysis. 
   
    \textbf{Likely Defective:} The lemma's alleged defective form occurs $\leq 10$ times in the corpus, indicating significant rarity, non-use, or absence.
  
    \textbf{On the Edge:} The lemma's alleged defective form occurs 11-100 times in the corpus.
  
    \textbf{Attested but Not Defective:} The lemma's forms occur frequently in the corpus, suggesting usage despite being listed as inflectional gaps in Wiktionary.
  

\begin{table}[h]
    \centering
    \begin{tabular}{lrr}
        \toprule
        Occurrences & \multicolumn{1}{c}{Latin} & \multicolumn{1}{c}{Italian} \\
        \midrule
        Likely defective:
        $\le$ 10    & 67.4\% & 79.2\% \\
        On the edge:
        11 - 100  & 25.4\% & 17.0\% \\
        Likely not defective:
        $>$ 100     & 7.2\% & 3.8\% \\
        \bottomrule
    \end{tabular}
    \caption{Validation of Wiktionary's defective verbs}
    \label{tab:defectivity}
\end{table}
\begin{table}[h]
    \centering
    \begin{tabular}{lrr}
        \toprule
        Log-Odds Ratio & \multicolumn{1}{c}{Latin} & \multicolumn{1}{c}{Italian} \\
        \midrule
        $>$ 1.9 & 6.3\%  & 0.0\% \\
        $>$ 1.5 & 12.2\%   & 5.9\% \\
        \bottomrule
    \end{tabular}
    \caption{Verbs found to be likely non-defective due to very high $p_w$ relative to $p_l \cdot p_f$}
    \label{tab:non_defective_verbs}
    
\end{table}

Based on the results shown in Table 1, Wiktionary's list of defective verbs in Latin is 1.8 times more likely to contain errors compared to Italian. This may be due to (1) the larger number of contemporary Italian speakers, leading to a stronger collective understanding of the language, and (2) Italian's less complex inflectional system compared to Latin. Table 2 shows the percentages of purported defective verbs that appear very frequently, relative to expected frequency. Following prior work, the Log-Odds Ratio threshold of large divergence is set at 1.9 \cite{Chen2010, Cohen2013}. Based on the Log-Odds Ratio model, approximately 6.3\% of Latin lemmata labeled as defective in Wiktionary may actually be non-defective. Similarly, the absolute frequency measure indicates that approximately 7\% of Wiktionary-listed defective Latin verbs are highly likely to be non-defective.
\subsection{Latin Results}

For Latin, 1,190 defective lemmata are sourced from Wiktionary. Of these, 1,050 lemmata (88\%) are attested in the corpus. Among the attested lemmata, 67\% exhibit defective behavior (i.e., some forms suggested by Wiktionary are verified to have extremely low frequencies). 
For example, \emph{discrepo} `to disagree' is a defective lemma. Wiktionary claims that \emph{discrepo} lacks a passive voice, and we found \emph{discrepo} to occur only 3 times in the passive voice. However, \emph{excommunico} `to excommunicate' is an example of Attested but Not Defective Lemmata as it is claimed by Wiktionary to lack a perfect aspect but actually has a perfect form that occurs 846 times. Examples of Not Attested Lemmata are \emph{astrifico} `to make stars', \emph{superfulgeo} `to shine', and \emph{auroresco} `to dawn'. 
\subsection{Italian Results}

For Italian, 124 defective lemmata are obtained from Wiktionary, and 103 (83\%) are attested in the corpus. Of the attested lemmata, $79\%$ exhibit defective behavior. For example, \emph{vèrtere} `to concern' occurs 6 times in the past participle form, below the threshold of 10, corroborating Wiktionary's claim that \emph{vèrtere} has no past participle form. 

Our system identifies potential candidates for errors in Wiktionary, such as \emph{consumere} `to consume', \emph{concernere} `to concern', and \emph{malandare} `to be ruined'. In some respects, it is understandable that the system flags these as unlikely to be defective even though they may actually be defective. For example, some native speakers confuse \emph{consumere} with \emph{consumare} `to consume' (sometimes mistakenly perceiving the word as a more formal variant), despite the fact that \emph{consumere} does not exist in modern Italian and is instead an archaic remnant from Latin. \emph{ludendo} `playing' is another word detected by our model to be unlikely to be defective. However, \emph{ludendo} appears frequently in the corpus due to code-switching with Latin.

Several verbs are identified by UDTube as past participle forms of allegedly defective verbs that lack a past participle form, such as \emph{lucido} and \emph{malandato}. However, these words are actually used adjectivally, meaning there is no contradiction with Wiktionary in these cases. 



\section{Conclusion}
This study presents a novel computational approach, based on cognitive modeling of language acquisition and the principle of Indirect Negative Evidence (INE), to validate defectivity in a widely used crowd-sourced linguistic resource. 
By integrating mBERT-enhanced UDTube with large-scale corpus analysis, we systematically evaluate the accuracy of defective verb classifications in Wiktionary. Our findings highlight the potential and limitations of crowd-sourced linguistic references while demonstrating the effectiveness of scalable NLP models, such as UDTube, in verifying morphological gaps in less-studied languages. The results indicate that Wiktionary is a reasonably reliable resource, with limitations. This study hence illustrates the importance of computational validation for crowd-sourced linguistic data as the results show that some verbs marked as defective in Wiktionary are, in fact, functional and widely used.
Additionally, the observed differences between Italian and Latin results suggest that linguistic evolution and corpus representativeness may impact the reliability of crowd-sourced morphological knowledge. Latin exhibits more inconsistencies, thus highlighting the need for careful interpretation of crowd-sourced knowledge and corpus-based evidence in the absence of native speakers. 


Future research can expand upon this work in several ways. First, the methodology here can be extended to other morphologically rich languages to assess the completeness and accuracy of crowd-sourced resources. Additionally, beyond defective verbs, this approach can be applied to other linguistic features, while integrating more diverse corpora could enhance our understanding of morphological productivity and gaps. Finally, improving neural morphological analyzers and experimenting with thresholds could enhance their ability to distinguish rare but valid forms from true gaps. 

By bridging computational methods with linguistic inquiry, the novel empirical results of our paper demonstrate how NLP and theories of language acquisition can enhance the quality assurance of crowd-sourced linguistic resources. The study also uniquely contributes to expanding linguistic databases and our understanding of language structure across typologically diverse systems. Finally, identifying and accounting for defectivity can improve the accuracy of morphologically aware NLP systems.

\section{Limitations}
Our use of mBERT may not yield the best performance, and future work could explore whether models like XLM-RoBERTa provide more accurate results for Latin and Italian. The corpora also have limitations, particularly in Latin, as certain verb forms may be underrepresented or entirely absent. Some rare or archaic forms may exist in texts outside the dataset, affecting the accuracy of frequency-based and statistical assessments. Additionally, context and pragmatics influence defectivity—some verbs classified as defective may still function within specific dialects, historical periods, or contexts.  Furthermore, since no standardized thresholds exist for determining defectivity, our criteria remain somewhat arbitrary. These limitations suggest that while corpus analysis provides valuable insights into the functional status of defective verbs, it should be supplemented with qualitative linguistic expertise and historical context. 

Another way that results may be impacted is the accuracy of UDTube. As expected from any models, UDTube is not perfect. Acknowledging that the annotation of morphological characteristics (FEATS) remains challenging, we chose UDTube due to its demonstrated superior performance in comparative studies \cite{yakubov2024how}. Our tuned UDTube model achieved 96\% accuracy on the Italian holdout test set and 98\% accuracy on the Latin holdout test set. Future work may further measure the performance of morphological analyzers in recent shared tasks, such as EvaLatin \cite{sprugnoli2022evalatin}, to advance evaluation standards for morphological analysis.
Additionally, as annotating the corpora is a computationally intensive task, we used distributed computing to complete the tagging in a reasonable timespan. Along the way, some nodes failed to complete their task, leaving some parts of the corpora untagged. Some cases of the limitations addressed above may have been avoided had the remaining portion of the corpora been used, but this is likely insignificant. 

\nocite{*}
\bibliography{custom}
\newpage
\appendix
\section{Paper Checklist}
\subsection{Benefits}
\begin{enumerate}
\item How does this work support the Wikimedia community?

This work helps enhance the quality and credibility of Wiktionary as a linguistic resource, particularly for under-documented or morphologically complex languages. It also provides replicable computational methods with tools that could be extended to other Wikimedia data to improve language data consistency and usability for linguistic and NLP applications. We computationally did quality assurance for the reliability of Wiktionary’s crowd-sourced morphological data in two languages. By applying a computational approach grounded in cognitive modeling, it identifies which defective verb entries in Wiktionary align with empirical corpus data and which may require more verification. 

\item What license are you using for your data, code, and models? Are they available for community re-use?

UDTube, a novel morphological analyzer on which this work builds, is released under the Apache 2.0 license, It is available on Github for community re-use. We also provide support for the tool.

\item Did you provide clear descriptions and rationale for any filtering that you applied to your data?

Yes. This study focuses on Latin and Italian, selected due to their morphologically rich systems and the availability of sizeable Wiktionary lists for defective verbs in both languages. We also used only morphological data on these languages on Wiktionary. 
\end{enumerate}

\subsection{Risks}
\begin{enumerate}
    \item If there are risks from your work, do any of them apply specifically to Wikimedia editors or the projects?
    
    This work does not pose direct risks to Wikimedia editors or the broader projects. It does not analyze or attribute data to specific contributors.

\item Did you name any Wikimedia editors (including username) or provide information exposing an editor's identity?

No. The study exclusively analyzes the linguistic content in Wiktionary. No usernames or editor identities were extracted, referenced, or disclosed.

\item Could your research be used to infer sensitive data about individual editors?

No. The research operates at the level of content (e.g. morphological content) and aggregate percentages and metrics. It does not involve editor metadata or revision history. There is no risk of inferring sensitive information about individual contributors.
\end{enumerate}

\end{document}